\documentclass[letterpaper]{article} % DO NOT CHANGE THIS
\usepackage{aaai22}  % DO NOT CHANGE THIS
\usepackage{times}  % DO NOT CHANGE THIS
\usepackage{helvet}  % DO NOT CHANGE THIS
\usepackage{courier}  % DO NOT CHANGE THIS
\usepackage[hyphens]{url}  % DO NOT CHANGE THIS
\usepackage{graphicx} % DO NOT CHANGE THIS
\usepackage{natbib}  % DO NOT CHANGE THIS AND DO NOT ADD ANY OPTIONS TO IT
\usepackage{caption} % DO NOT CHANGE THIS AND DO NOT ADD ANY OPTIONS TO IT
\usepackage{booktabs}
\usepackage{multirow}
\usepackage{amsmath}
\usepackage{color}
\DeclareCaptionStyle{ruled}{labelfont=normalfont,labelsep=colon,strut=off} % DO NOT CHANGE THIS
\usepackage{xcolor}

\usepackage{newfloat}
\usepackage{algorithm}
\usepackage{algorithmic}

\usepackage{newfloat}
\usepackage{listings}
\floatstyle{ruled}
\newfloat{listing}{tb}{lst}{}
\floatname{listing}{Listing}
\title{Reliability Analysis of Psychological Concept Extraction and Classification in User-penned Text}
\title{Reliability Analysis of Psychological Concept Extraction and Classification in User-penned Text}
\title{Reliability Analysis of Psychological Concept Extraction and Classification in User-penned Text}
\author {
    Muskan Garg, \textsuperscript{\rm 1}
    MSVPJ Sathvik, \textsuperscript{\rm 2}
    Amrit Chadha, \textsuperscript{\rm 3}
    Shaina Raza, \textsuperscript{\rm 4}
    Sunghwan Sohn \textsuperscript{\rm 1} 
}
\affiliations {
    \textsuperscript{\rm 1} Mayo Clinic, Rochester, MN, USA, 
    \textsuperscript{\rm 2} IIIT Dharwad, Karnataka, India, \\
    \textsuperscript{\rm 3} IIT Kharagpur, India,
    \textsuperscript{\rm 4} Vector Institute, Ontario, Canada
    
    garg.muskan@mayo.edu, 20bec024@iiitdwd.ac.in, amritchadha66@gmail.com,\\ shainaraza@vectorinstitute.ai, sohn.sunghwan@mayo.edu
}
\usepackage{bibentry}
\begin{document}

\maketitle

\begin{abstract}
The social NLP research community witness a recent surge in the computational advancements of mental health analysis to build responsible AI models for a complex interplay between language use and self-perception. Such responsible AI models aid in quantifying the psychological concepts from user-penned texts on social media. On thinking beyond the low-level (\textit{classification}) task, we advance the existing binary classification dataset, towards a higher-level task of reliability analysis through the lens of explanations, posing it as one of the safety measures. We annotate the \textit{LoST} dataset to capture nuanced textual cues that suggest the presence of low self-esteem in the posts of Reddit users. We further state that the NLP models developed for determining the presence of low self-esteem, focus more on three types of textual cues: (i) \textit{Trigger}: words that triggers mental disturbance, (ii) \textit{LoST indicators}: text indicators emphasizing low self-esteem, and (iii) \textit{Consequences}: words describing the consequences of mental disturbance. We implement existing classifiers to examine the attention mechanism in pre-trained language models (PLMs) for a domain-specific psychology-grounded task. Our findings suggest the need of shifting the focus of PLMs from \textit{Trigger} and \textit{Consequences} to a more comprehensive explanation, emphasizing \textit{LoST indicators} while determining low self-esteem in Reddit posts. %We release this dataset as the second version of LoST dataset as \textsc{LoST.v2} at Github.
\end{abstract}

%% main text
\section{Background}

% \textcolor{red}{\textbf{CONNECT}}
Mental disorders are a significant contributor to global mortality rates, accounting for approximately 14.3\% of deaths worldwide~\cite{walker2015mortality}. According to \textit{the Global Burden of Diseases, Injuries, and Risk Factors Study 2019}, mental disorders continue to rank among the top ten causes of burden globally, without any indication of a reduction in their impact since 1990~\cite{gbd2022global}.
In the past few years, extensive investigations have illuminated the intricate associations between low self-esteem and various mental disorders, as evidenced by the Diagnostic \& Statistical Manual of Mental Disorders (DSM-5)~\cite{rouault2022low, cella2022non}. 
\begin{figure}[ht]
    \centering
    \includegraphics[width=0.47\textwidth]{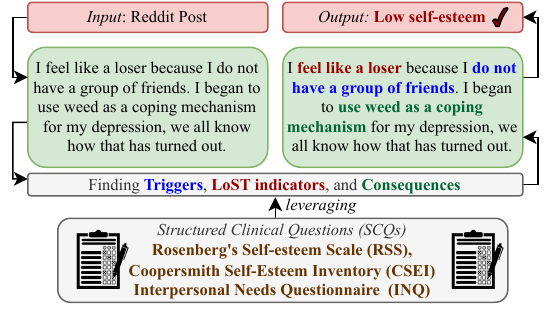}
    \caption{The overview of our task. We annotate the textual cues indicating the low self-esteem aspect in human-writings, emphasizing the need of focusing LoST indicators more than triggering words and final consequences.}
    \label{fig1}
\end{figure}
\paragraph{Psychology-Ground.}
The notion of \textit{low self-esteem} has been identified as a predisposing factor for the onset of social anxiety~\cite{acarturk2009incidence}. In line with this, scholarly evidence underscores the pivotal role of low self-esteem in heightening the susceptibility to depression, anxiety, suicidal tendencies, impaired cognitive performance, compromised sleep quality, and diminished overall health~\cite{korkmaz2019suicide}. Additionally, young individuals with recurrent suicide attempts exhibit a pronounced tendency towards enduring suicidal thoughts, challenges in interpersonal relationships, feelings of detachment, and diminished self-efficacy~\cite{choi2013risk}. Historical studies have shed light on the diminished self-assurance prevalent among individuals with low self-esteem, and its consequential association with reduced social involvement~\cite{watson2012rejection}. Emphasizing the paramountcy of \textit{high self-esteem} as a foundational human requisite, a distinguished American psychologist has delineated two distinct variants of ``esteem'' within his seminal theory on the \textit{`hierarchy of human needs'}: (i) the \textit{external validation from peers}, encompassing acknowledgment, accomplishments, and respect, and (ii) the \textit{intrinsic self-regard}, encapsulating self-affection, assurance, capability, and prowess~\cite{cox1987rich}.  

%According to Cox's theory of the "hierarchy of human needs", having a strong sense of self-worth is considered a primary requirement~\cite{cox1987rich}. Cox differentiates between two forms of "esteem": the first being the "need for recognition, success, and admiration from others," and the second the "need for self-respect, self-confidence, and proficiency." 
The correlation between low self-esteem and 21 distinct disorders is well-documented, encompassing diagnostic criteria, associated features, risk factors, and consequences specifically linked to individuals grappling with low self-esteem~\cite{kolubinski2018systematic}. 
%The evidences of low self-esteem are hidden in the text indicators when an individual express his thoughts in self-narrated text
% \textcolor{red}{\textbf{CONNECT}}
Individuals often share their feelings and thoughts on social media, especially when reflecting on personal experiences related to low self-confidence. Identifying and understanding the deep connection between low self-esteem and related disorders is crucial for developing effective interventions and promoting mental health.

% Hence

\begin{table*}[h]
 \small
    \centering
    \begin{tabular}{p{0.7cm}|p{14.5cm}|p{1.1cm}}
      \toprule[1.5pt]
        \textbf{S.No.}& \textbf{Text} & \textbf{Label}\\
        \midrule
        T1 & ... now i was diagnosed with mental illness and he is disappointed because he is afraid I will turn like my mother. Yesterday at night I went to the bathroom which is next to my parents bedroom and he said to my mom about her but then he also included me ''\textcolor{blue}{i am disgusted by fat}$\leftarrow$ (\texttt{trigger}), its \textcolor{blue}{disgusted, filthy and dirty}$\leftarrow$ (\texttt{trigger}) i don't even know how to express it to my daughter''. It just \textcolor{blue}{broke me}$\leftarrow$ (\texttt{trigger}), broke me to pieces. \textcolor{red}{I doubt he even loves me}$\leftarrow$ (\texttt{LoST}), i have seen him looking at me with \textcolor{blue}{disappointment}$\leftarrow$ (\texttt{trigger}) when i eat something it's just makes me \textcolor{red}{hate my self more}$\leftarrow$ (\texttt{LoST}) and make me \textsc{want to kill myself}$\leftarrow$ (\texttt{consequences})... & Presence\\
        \midrule
        T2& I'm \textcolor{red}{boring}$\leftarrow$ (\texttt{LoST}). I'm \textcolor{red}{not good}$\leftarrow$ (\texttt{LoST}) at socializing and I'm very \textcolor{red}{awkward}$\leftarrow$ (\texttt{LoST}). I'm \textcolor{red}{replaceable}$\leftarrow$ (\texttt{LoST}). **** I \textsc{want to die}$\leftarrow$ (\texttt{consequences}) so much right now. & Presence \\
        \midrule
        T3& It's night and as usual that's when all these \textcolor{blue}{horrible thoughts come to my head}$\leftarrow$ (\texttt{trigger}). I know I'm \textbf{not a terrible person} but I haven't exactly had the best year so I feel kind of overwhelmed right now. I really need to talk to someone ASAP \textsc{I don't feel good}$\leftarrow$ (\texttt{consequences}). & Absence\\
        \midrule
        T4 & So, I have been thinking of \textsc{harming myself}$\leftarrow$ (\texttt{consequences}) lately even though it's been a while since I've done it. %I've been able to keep myself from harming myself or doing drugs for a while now but something just snapped in recent months that's making me just irritable, impatient, and wanting to spiral out of control again. I just feel like a shell and any kind of physical pain is like breath of fresh air. Anyone else go through something like this? If you got through it, how?
        & Absence\\
        \toprule[1.5pt]
    \end{tabular}
    \caption{Samples of the dataset. We label a given text along with three textual cues (i) triggering (blue colored text), (ii) LoST indicators (red colored text) and (iii) consequences (capitalized text).}
    \label{tab:samples}
\end{table*}

% \paragraph{Clinical Questionnaires.}

 \paragraph{Reliability Analysis: Motivation.}
In the nuanced arena of analyzing user-penned content on social media for indicators of low self-esteem, reliability is paramount. As mental health practitioners and platforms increasingly recognize the value of insights from social media to tailor interventions, the role of trustworthy models becomes even more critical. Thus, the intersection of mental well-being and technological analysis demands both accuracy and trustworthiness. Reliable analysis ensures that the textual cues being flagged as indicative of low self-esteem are genuine, eliminating false positives and ensuring that genuine cases don't go unnoticed. By reliably interpreting an individual's online expression, there's an opportunity to provide more personalized guidance and resources. For such analytical models to be wholeheartedly embraced by both the professional community and the users themselves, they must be deemed reliable. % When a model consistently and accurately identifies text-spans that reflect low self-esteem, it engenders trust. This trust ensures that the insights and interventions based on these models are respected and acted upon. Moreover, these models, with their capacity to process vast amounts of social media content, are also invaluable for broader research endeavors. By reliably tracking the prevalence of low self-esteem context in user content, they offer a lens into societal mental health trends, which can, in turn, influence awareness campaigns and larger mental health initiatives.

\paragraph{Our Contributions.} 
The current landscape offers abundant open-source datasets for research and application development across various fields. Yet, there exists a notable dearth of high-caliber psychology-grounded datasets, especially pronounced in the ever-evolving realm of healthcare informatics and NLP.%Within this domain, datasets are indispensable for both training and performance assessment of Language Models. 
Datasets sourced from participatory healthcare informatics frequently exhibit attributes of intricacy, randomness, and an absence of clear structure. The performance of models is invariably tied with the integrity, and applicability of the dataset. With these considerations, a paramount challenge in psychology-driven dataset compilation is to ensure the consistency and reliability of the annotations. To navigate and simulate this intricate landscape, we choose to design an annotation scheme to discern textual-cues for psychological concept (low self-esteem) extraction and classification in user-penned text. %Such datasets, marked by their adaptability, underpin a multitude of tasks, from model training to evaluation. Yet, it's worth noting that t
The assembly process of our annotation scheme and annotated dataset reveals discernible gaps in content comprehensiveness and mark its adaptability for healthcare utilization.

Building upon this foundational knowledge, annotation guidelines were meticulously curated through the joint endeavors of two specialists: (i) a senior clinical psychologist and (ii) a social NLP expert.
%Our major contribution is to find textual cues indicating low self-esteem in a given text without making any prior assumptions about user's perspective. 
Our work enables the development of reliable NLP models in the near future, to facilitate \textit{higher-level tasks}~\cite{sheth2021knowledge}\cite{raza2023constructing}. As such, we contribute by advancing our studies %over the existing dataset, \textsc{LoST.v1}~\cite{garg2023lost}, and annotating it with %the \textit{textual-cues} %in all the positive samples\footnote{samples containing low self-esteem} with the lens of 
%with the psycholoconcept (low self-esteem) extraction and 
with the responsible classification of user-penned text through the lens of three types of textual cues: \textit{Trigger, LoST indicators, Consequences}, collectively we mention them as TLC. To this end, we define this task as the first of its kind to detect appropriate textual cues that aid %a strong focus on reliability analysis %\textit{enhancing the attention mechanism} 
in decision-making of  
low self-esteem detection. %The objective of this research paper is to find the textual-cues that indicate low self-esteem of Reddit users in their posts. 
%paving the way for developing more comprehensive, reliable and robust models. For this work, 
We define these textual cues as set of words indicating "explainability" and are focused by the attention mechanism of classifiers while making decisions.

 %such as personalization, contextualization and abstraction in mental health analysis leveraging fine-grained annotation of human-writings. For this work, we define these textual cues as set of words indicating "explainability" and are focused by the attention mechanism of classifiers while making decisions. 

% \paragraph{Problem Formulation} 

% For instance, consider the Table~\ref{tab:samples}.
\section{Corpus Construction}
\paragraph{Corpus Collection}
We construct a dataset for identifying text-spans reflecting low self-esteem, from the collected instances from subreddits \texttt{r/depression} and \texttt{r/suicidewatch} from 2 December 2021 to 4 January 2022. By using this curated dataset, we aim to ensure the quality and relevance of the data for our specific research objectives. We further perform the manual cleaning and removed all nearly empty posts (posts having less than 10 words) and posts containing only URLs to accommodate the subjectivity of the task. We included the user-penned experiences and excluded all the general posts borne out of concerns for fellow members of the subreddit community. We finally obtain $2174$ samples, and annotate textual cues in $465$  ($\approx$25\%) positively labeled samples for TLC (\textit{trigger}: $3988$ words; \textit{LoST indicators}: $6514$ words; \textit{consequences}: $787$ words). 

Tackling the intricate and highly subjective task of identifying low self-esteem in textual content can inadvertently lead to mistakes when relying solely on \textit{naive} judgment. To address this challenge, our approach encompassed forming a collaborative unit comprising a clinical psychologist (tasked with discerning the nuanced psychological undertones within the text) and a social NLP expert (responsible for meticulous text annotations conducive for advanced AI models). To harmoniously blend insights from clinical psychology with the NLP realm, domain experts propose granular guidelines that categorize low self-esteem as an underlying psychological concept marked by self-doubt, feelings of worthlessness, and a discernible absence of confidence. Aiding the annotation chore, our domain experts contrived a structured annotation blueprint anchored on clinical questionnaires and two pivotal research queries: (i) ``RQ1: \textit{Is the text posted in the subreddits related to depression and suicidal ideation shows the sign of low self-esteem through direct text-spans ?}", and (ii) ``RQ2: \textit{To what extent should annotators delve into the text to identifying text-spans for low self-esteem?}".

\paragraph{Annotation Scheme} When an individual experiences low self-esteem within the context of a mental health condition, healthcare professionals usually assign diagnostic codes that correspond to the specific mental health disorder contributing to or causing the low self-esteem. We define our annotation task as the identification of text-spans that encapsulate the essence of low self-esteem. 

Given the inherently subjective and inference-driven nature of textual data, our research aims to develop and disseminate comprehensive annotation guidelines for discerning clinical concepts within text and subsequently translating them into diagnostic codes. It's worth noting that within the International Classification of Diseases, 10th Revision (ICD-10), there isn't a dedicated code specifically designated for ``low self-esteem'' as a primary diagnosis. The ICD-10 primarily focuses on classifying medical conditions and diseases, with mental health conditions, including issues related to self-esteem, typically categorized within broader classifications like mood disorders or neurotic disorders. Some of the major ICD codes are: F32 - Major depressive disorder, Dysthymic disorder (Persistent depressive disorder), Panic disorder (with or without agoraphobia), Anxiety disorder, Obsessive-compulsive disorder, Reaction to severe stress, and adjustment disorders, Dissociative and conversion disorders, Somatoform disorders (which include somatic symptom disorders).

Our experimentation endeavors to establish the groundwork for automating the conversion process, enabling the transformation of text, whether it originates from user-generated content or clinical notes, into relevant diagnostic codes. We further utilize standard clinical questionnaires (SCQ) to frame annotation scheme to supervise the correct annotations. Within this framework, a collaboration between a clinical psychologist and a social NLP researcher has resulted in the creation of annotation guidelines based on structured clinical questions (SCQs). These SCQs encompass a collection of questions rooted in psychological principles, which are essential for extracting precise and pertinent information. Furthermore, these questions employ assessment tools to collect data and appraise a variety of psychological constructs. The foundation for our dataset creation lies in the SCQs derived from three primary clinical surveys: (i) Rosenberg's Self-esteem Scale (RSS)~\cite{rosenberg1965rosenberg}, (ii) Coopersmith Self-Esteem Inventory (CSEI)~\cite{potard2017self}, and (iii) Interpersonal Needs Questionnaire (INQ-18)~\cite{mitchell2020interpersonal} (see Figure~\ref{fig1}). We take annotations to identify three types of text-spans in a given text: Triggering, LoST, and Consequences (TLC).
% at two levels:
% \begin{itemize}
%     \item We first 
%     \item We classify the text as $1$ if the text-spans in LoST is not null, else $0$.
% \end{itemize}
We define TLC as follows:
\begin{enumerate}
    \item \textbf{Triggering}: The reason behind low-self esteem is a triggering component that further enhances differentiation between the \textit{low self-esteem} and an \textit{event that may incur low self-esteem} in a person. We instruct the annotators to identify text-spans causing mental disturbance.
    \item \textbf{LoST}: The text spans that indicates one of the 10 pre-defined annotation principles in the first-person context, should be identified as low self-esteem. 
    For example, there is a substantial difference between \texttt{my friends think I am not funny} and \texttt{I am not funny}. Although the text-span \texttt{I am not funny} is present in both these statements, however, the former one should be marked as absence of low self-esteem due to public opinion which may be a seed to implant the prospective low self-esteem. However, the cross-sectional study cannot reveal users' perception about public opinion and hence, we do not make any assumptions.
    \item \textbf{Consequences}: %To differentiate the consequences from low self-esteem, we find the text-spans indicating one or more predefined principles in annotation scheme. 
    There is a substantial difference between text indicators of final state of mental disturbance (self-harm and suicidal ideation), and low self-esteem. We instruct the annotators to identify text-spans casting signs of severe consequences due to mental disturbance such as \textit{'feels like the end is near', 'feeling trapped'}.
\end{enumerate}

% \paragraph{Annotations}
Our experts developed the annotation guidelines for identifying textual cues team leveraging SCQs. Three postgraduate students carry out the manual annotations after successfully completing experts-driven training sessions. Annotations were carried out using the expert-driven annotation scheme, designed to ensure \textbf{consistency} and \textbf{synchronization} during annotations. 
%We ensure their \textbf{correctness} for alignment and understanding of the task to facilitate coherent annotations. Due to the complex nature of the task, we found a slight lack of synchronization among annotators. 
%To resolve this problem, w

Annotators were made to sit together and annotate the text spans for TLC, resulting in \textit{one group-annotation} to facilitate coherent annotations. This annotation task was followed by experts' validation. 

\paragraph{Inter-Annotator Agreement} After experts' validation, we test the reliability of their judgement for all the $465$ samples using \textit{Fliess' Kappa inter-observer agreement} study~\cite{guggenmoos1996meaning}. We employ two experts to verify the annotations by marking them as either \textit{acceptable} or \textit{unacceptable}. To quantify the agreement, we calculate $\kappa$ for \textit{acceptance} and notice $67.52\%$, $71.92\%$, and $69.32\%$ of agreement among annotators for \textit{Trigger, LoST indicators, and Consequences}, respectively. We acknowledge that the lower value of inter-annotator agreement are a well-known problem in emotion-based subjective studies, where lower agreement scores are reported~\cite{tsakalidis2018building}.  %We obtain final annotations based on the \textit{majority voting} mechanism. 
The samples of our dataset in Table~\ref{tab:samples}, exemplifies the concept of explainable low self-esteem detection. We observe that all three types of textual cues in TLC, are present in $T1$ but \texttt{trigger} is missing in $T2$. We illustrate that the presence of \texttt{consequences} does not ensure the presence of \texttt{LoST indicators} (see $T3$ in Table~\ref{tab:samples}).

\paragraph{FAIR Principles} The FAIR principle~\cite{wilkinson2016fair}, enhances the findability, accessibility, interoperability, and reusability of datasets to emphasize the actionable nature of machine-centric systems, which are becoming increasingly relied upon in facilitating future research endeavors. Our dataset contains the [\textit{text}, \textit{label}, \textit{Trigger}, \textit{LoST indicators}, and \textit{Consequences}] for all the positively labeled data-points and later three labels (TLC) for the other negatively labeled data-points. The dataset is constructed in the comma-separated format\footnote{The dataset shall be made available on request via signed agreement.}. We plan to expand and update our dataset with more data-points and more aspects of mental health in upcoming versions. %such as explainable loneliness detection~\cite{garg2023lonxplain}. 
%We sufficiently annotate the \textsc{LoST.v2} for an explainable binary classification problem to facilitate its re-usability. In order to uphold the ethical principles of privacy, safety, and accountability, we have abstained from disclosing any metadata in the public domain. %However, our dataset remains a valuable resource that can be effectively utilized to explore novel research avenues in the domain of NLP-centered mental health analysis. 
By respecting the necessary safeguards, we aim to ensure the responsible use of this dataset while enabling advancements in understanding different aspects of mental health through computational linguistic approaches. 
In future, we plan to enhance the other aspects of our sister datasets such as explainable loneliness detection~\cite{garg2023lonxplain} on the same lines. %, encouraging the development of improved attention mechanisms for contextual-PLMs.

\paragraph{Inferences}
%We acknowledge that the textual cues in ground-truth are extracted from the original text leveraging experts-driven annotation scheme. % Three different types of textual cues suggest the (i) cause (triggering words), (ii) , and (iii) consequence of the mental disturbance along with a specific aspect modeling such as \textit{LoST indicators}. Furthermore, w
We acknowledge that only 33.12\% of the positively labeled data-points contains all three types of textual cues among TLC. We find $465$ $textual-cues$ for LoST indicators, overlapping with \textit{Trigger} and \textit{consequences} upto 85.16\% and 40.64\%, respectively. The natural composition of the dataset is $\approx$$\frac{3}{1}$ ratio for negative (0) to positive (1) label where positive (1) label indicates the presence of low self-esteem. The schema of our dataset is as follows:

\begin{quote}
% \small
    \textit{$<$ \textbf{Text} (string), \textbf{Label} (binary)$>$, \\ \textcolor{blue}{// Text-classification problem}\\ \\$<$ \textbf{Trigger} (list of string),\\\textbf{LoST indicators} (list of string), \\\textbf{Consequences} (list of string)$>$ \\\textcolor{blue}{//  Reliability analysis}}
\end{quote}

Furthermore, we examine $T3$ in Table~\ref{tab:samples}, where the individual refers themselves as \texttt{`not a terrible person'}, indicating that the person does not possess any thoughts of low self-esteem. However, it's worth noting that words such as \texttt{`I'}, \texttt{`terrible'}, and \texttt{`person'} may not provide reliable predictions, resulting in negative perception of the user. To this end, we highlight the importance of considering semantic enhancements to develop more comprehensive and informative models. 

\section{Proposed Method}
The idea behind reliability analysis is to identify text-spans that are aimed at detecting the presence of low self-esteem in user-penned content on social media. We apply \textit{BERT}, a model tailored to detect text-spans in user-penned posts that potentially signify low self-esteem. The model systematically processes posts from a set \( P \) and classifies them accordingly.

\subsection{Problem Formulation}
\begin{figure*}
    \centering
    \includegraphics[width=0.92\textwidth]{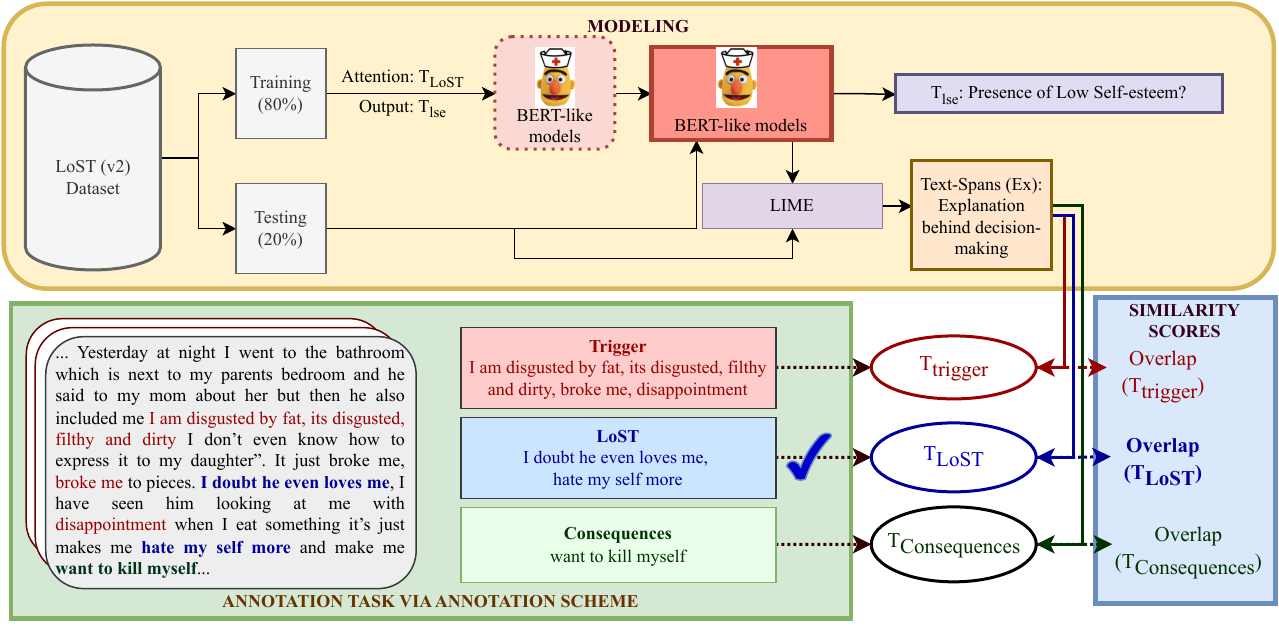}
    \caption{Architecture of Reliability Analysis for Low self-esteem detection and classification in user-penned text.}
    \label{fig2}
\end{figure*}

Given a collection of posts, represented as \( P = \{P_1, P_2, \ldots, P_n\} \), our model operates in two primary phases: \textit{an attention mechanism} to discern the relevant text-spans and \textit{a classification mechanism} to decide the overall sentiment of the post concerning self-esteem as shown in Figure~\ref{fig2}.
\paragraph{Attention Mechanism}
The core of pre-trained language models is the attention mechanism that dynamically computes weights, or "attention scores", for different parts of the input, allowing the model to focus selectively on specific parts of the input, especially areas that may signal low self-esteem. Given a post \( P_i \), each token in a post has an associated attention weight. Formally, the attention mechanism \( A \) for post \( P_i \) can be represented as:
\begin{equation}
A(P_i) = \{a_1, a_2, \ldots, a_m\}
\end{equation}
Where:
\begin{itemize}
    \item \( a_j \) stands for the attention weight assigned to the \( j^{th} \) token of post \( P_i \).
    \item \( m \) signifies the total number of tokens within post \( P_i \).
\end{itemize}

% In practice, the attention weights are determined using sophisticated models, potentially employing scaled dot-product attention or multi-head attention techniques, among others.

\paragraph{Classification Mechanism}
Once we compute the attention weights, the subsequent step involves classifying the text based on its content. The classification leans heavily on the attention scores, using them in combination with the embedded representations of the tokens to arrive at a binary decision. The function \( C \) represents this classification step. Given the attention weights from function \( A \) and the token embeddings, \( C \) produces an outcome indicating if the post manifests signs of low self-esteem:
\begin{equation}
C(A(P_i)) = 
\begin{cases} 
    1 & \text{if low self-esteem is detected} \\
    0 & \text{otherwise}
\end{cases}
\end{equation}

In essence, for every post \( P_i \) in our set \( P \), the model follows a two-step process:
\begin{enumerate}
    \item Compute attention weights using the function \( A \).
    \item Use these weights, along with the post's token embeddings, to classify the post with the function \( C \). The result denotes the detected sentiment: \( 1 \) for low self-esteem and \( 0 \) otherwise.
\end{enumerate}

\subsection{Mathematical Notations}

The core utility of attention mechanisms in NLP models, such as BERT, is to ascertain the significance of individual tokens or spans within larger sequences. For our objective of \textit{\textbf{identifying indicators of low self-esteem}} in user-penned text, we aim to emphasize specific words or phrases that may suggest such clinical concepts.

\begin{enumerate}
    \item \textit{Token Embeddings:} Every token \( t \) within a post \( P_i \) is mapped to a dense vector, denoted as \( E(t) \). This embedding captures the semantic nuance of the tokens, offering a representation in higher-dimensional space.

    \item \textit{Transformations to Query, Key, and Value Vectors:} Each token is further transformed to generate its query, key, and value vectors. These are critical for determining the relative importance of a token in the context of the entire post. 

    \begin{align*}
    Q(t) &= E(t) W_Q \\
    K(t) &= E(t) W_K \\
    V(t) &= E(t) W_V
    \end{align*}

    Where: \( W_Q, W_K, \) and \( W_V \) are trainable weight matrices for the query, key, and value transformations respectively.

    \item \textit{Attention Score Computation:} For each token, an attention score \( s_j \) is derived to measure its relevance with respect to every other token, particularly in the context of detecting signs of low self-esteem:

    \[
    s_j = Q(t_j) \cdot K(t)^T
    \]

    \item \textit{Normalization via Softmax:} The raw scores are normalized using the softmax function to make them lie between 0 and 1, ensuring they sum up to 1:

    \[
    \alpha_j = \frac{\exp(s_j)}{\sum_{k=1}^m \exp(s_k)}
    \]

    Here, \( \alpha_j \) is of utmost importance. Tokens with particularly elevated \( \alpha_j \) values are ones that the model considers critical in signifying low self-esteem.
\end{enumerate}

% \subsubsection{Classification Mechanism: Pinpointing Low Self-Esteem}

Upon computing the attention weights, our subsequent goal is to use this enriched data to \textit{\textbf{classify}} the conceptual significance of the text, specifically, determining if the emphasized tokens suggest low self-esteem.

\begin{enumerate}
    \item \textit{Collated Post Representation:} Considering the varying attention weights for each token, we generate a consolidated representation for the post \( P_i \):

    \[
    R(P_i) = \sum_{j=1}^m \alpha_j A(t_j)
    \]

    \item \textit{Through the Classification Layer:} This combined vector is then pushed through a dense layer which outputs a probability score \( p \). This score represents the likelihood that the post exhibits low self-esteem:

    \[
    p = \sigma(R(P_i) W_C + b_C)
    \]

    Where:
    \begin{itemize}
        \item \( W_C \) is the weight matrix associated with the classification layer.
        \item \( b_C \) is the bias term.
    \end{itemize}

    \item \textit{Binary Classification Based on Threshold:} Depending on the value of \( p \), a binary classification decision is made:

    \[
    C(A(P_i)) = 
    \begin{cases} 
        1 & \text{if } p \geq 0.5 \text{ (indicative of low self-esteem)} \\
        0 & \text{otherwise}
    \end{cases}
    \]
\end{enumerate}

% \section{Loss Function: Binary Cross-Entropy Loss}

For \textit{\textbf{binary classification tasks}}, one of the most commonly employed loss functions is the binary cross-entropy loss. Given the predicted probability \( p \) and the true label \( y \) (where \( y = 1 \) indicates low self-esteem and \( y = 0 \) otherwise), the binary cross-entropy loss, \( L \), is defined as:

\[
L(p, y) = - \left( y \log(p) + (1 - y) \log(1 - p) \right)
\]

Where:
\begin{itemize}
    \item \( p \) is the predicted probability of the post being indicative of low self-esteem.
    \item \( y \) is the ground truth label.
\end{itemize}

\subsection{Reliability Analysis}

In our research, we delve deeply into the specific text regions, commonly referred to as ``text-spans'', that the BERT model directs its attention towards, in an attempt to identify indicators of low self-esteem within written content. As per our established methodology, each piece of text \( T \) is meticulously annotated into several categories:

\begin{equation}
T = \{T_{lse}, T_{trigger}, T_{LoST}, T_{consequences}\}
\end{equation}

where:
\begin{itemize}
    \item \( T_{lse} \) denotes the presence or absence of low self-esteem.
    \item \( T_{trigger} \) represents textual cues or triggers leading to potential mental disturbances.
    \item \( T_{LoST} \) indicates the textual span highlighting the author's perceived low self-esteem.
    \item \( T_{consequences} \) captures the aftermath or resulting mental state from the disturbances.
\end{itemize}

Given the model's attention mechanism, for any token \( t_i \) in \( T \), the attention weight is denoted as \( A(t_i) \). Our hypothesis posits that the model's attention, when detecting low self-esteem, is primarily distributed among the tokens related to the three categories of trigger, LoST, and consequences. Mathematically, the attention distribution for each category is:

\begin{equation}
A_{category} = \sum_{t_i \in T_{category}} A(t_i) + \epsilon
\end{equation}

where $\epsilon$ is the attention given to the words other than the ones in TLC category.
Ideally, the attention weights should be skewed towards the LoST category. Therefore, an optimal model's attention distribution would satisfy:

\begin{equation}
A_{LoST} > A_{trigger}
\end{equation}
\begin{equation}
A_{LoST} > A_{consequences}
\end{equation}

To evaluate the model's accuracy in focusing on the correct text-spans, we utilize an \textit{exact match algorithm}. For each category \( category \) in the text \( T \), the overlap with the model's explanations \( Ex \) is computed as shown in Equation~\ref{eqn7}:

\begin{equation}
\label{eqn7}
Overlap_{category} = \frac{|Ex \cap T_{category}|}{|T_{category}|}
\end{equation}

where \(  category \) belongs to a set containing all three categories: $\{  T_{trigger}, T_{LoST}, T_{consequences} \}$. The overlap gives a percentage indication of how well the model's attention or explanation aligns with the pre-annotated TLC text-spans.

\section{Experiments and Evaluation}

\subsection{Classifiers as baselines}
We on several models originating from the BERT architecture, each bringing its unique features to the table, specifically for concept extraction and classification.
\begin{itemize}
    \item \textbf{BERT} is a transformer-based architecture that captures contextual information from both left and right sides of a token in any input text~\cite{devlin2018bert}. This architecture has set a new standard for a range of NLP tasks. The bi-directionality of BERT ensures that each word is analyzed in its surrounding context, making it potent for discerning subtle cues indicative of low self-esteem, which can be context-dependent.
    \item \textbf{ALBERT} is a variant of BERT optimized for faster training and less memory consumption without compromising on performance~\cite{lan2019albert}. ALBERT's parameter-reduction techniques ensure that we maintain the power of BERT while achieving faster model training, which is beneficial when working with large and nuanced datasets, such as those involving human emotions and states.
    \item \textbf{DistilBERT} is a distilled version of BERT, retaining most of its performance capabilities but being 40\% smaller and 60\% faster~\cite{sanh2019distilbert}. Given the need for real-time or faster processing in social media content analysis, DistilBERT's speed and size advantages make it an attractive choice for extracting low self-esteem indicators without significant lag.
    \item \textbf{DeBERTa} improves upon BERT by disentangling the inter-token semantic relations with absolute positional encoding~\cite{he2020deberta}. The disentangled attention mechanism can be pivotal in decoding intricate user emotions and sentiments, ensuring that the model is sensitive to both the semantics and position of tokens when identifying low self-esteem cues.
    \item \textbf{ClinicalBERT}, as its name suggests, is fine-tuned on clinical narratives or medical literature, making it adept at understanding medical terminologies and contexts~\cite{huang2019clinicalbert}. When considering low self-esteem within a clinical or medical paradigm, ClinicalBERT's expertise can ensure that medical or clinical references related to self-esteem are accurately captured and classified.
    \item Being optimized for psychological contexts, \textbf{PsychBERT} is inherently more sensitive and attuned to detecting subtle emotional indicators, such as those found in content penned by individuals with low self-esteem~\cite{vajre2021psychbert}.
    \item Addressing the wider spectrum of mental health, \textbf{MentalBERT} offers a holistic approach to detect low self-esteem, considering it alongside other potential mental health indicators~\cite{ji2022mentalbert}.
\end{itemize}

\paragraph{Experimental Setup}
We implement the existing classifiers to test the accuracy and robustness of the models for identifying textual cues projecting low self-esteem. We split our dataset into a ratio of 80:20, with 80\% of the data allocated for training samples and 20\% reserved for testing purposes. The randomised distribution of training to testing data contains 376 positive samples out of 1,739 for training data, and 89 positive samples out of 435 samples. We use the validation set from training samples to fine-tune model hyperparameters and assess the performance during the training process. Among the classifiers, we consider four well-established pre-trained language models (PLMs) (BERT, ALBERT, DistilBERT, and DeBERTa), while the remaining three are domain-specific PLMs having strong association with mental health domain. We use the standard PLMs available on huggingface and default attention mechanism to perform experiments with existing classifiers. We set the learning rate (lr) as $2e-5$ for a batch size of $8$, weight decay as $0.01$, warmup steps of $100$, and logging steps of $100$ for fine-tuning PLMs. Furthermore, we obtain a new testing dataset of 200 samples from out-of-distribution (OOD) dataset, Dreddit, a publicly available data~\cite{turcan2019dreaddit}, originally constructed to classify depression and suicide risk. We annotate first 200 instances of Dreddit dataset for reliability analysis using the same annotation scheme. We use the Google colab pro environment for experiments and evaluation to access faster GPUs like the Tesla P100, and occasionally the Tesla T4 and Tesla V100. This was beneficial for running compute-intensive tasks. % The number of epochs and other hyperparameters are appropriately adapted while fine-tuning different models. 

\paragraph{Evaluation metrics} In this research paper, we employ \textit{precision, recall, F-score, accuracy, }and \textit{Matthew's correlation coefficient (MCC)~\cite{boughorbel2017optimal}} as evaluation metrics for identifying Reddit posts casting low self-esteem. MCC takes into account true and false positives and negatives and is generally regarded as a balanced measure, even when classes are of very different sizes. Additionally, we use LIME~\cite{ribeiro2016should} that generates faithful local explanations, to find system-level explanations and compare the resulting explanations with the ground-truth textual cues for (i) triggers, (ii) LoST indicators, and (iii) consequences, using two text similarity matching mechanisms~\cite{yang2018adaptations}: (i) Recall Oriented Understudy for Gisting Evaluation (ROUGE) scores, and (ii) BiLingual Evaluation Understudy (BLEU) scores. ROUGE provides several measures to determine the quality of a summary by comparing it with other (reference) summaries. BLEU compares the n-grams in the machine-generated translations to those in the reference translations and returns a score between 0 and 1, where 1 means the generated translation matches the reference perfectly. 

% \paragraph{Evaluation Metrics}

% \subsection{Experimental Results and Discussion}
% \textcolor{blue}{Could you please highlight the best results by making them bold? We compare the Accuracy and MCC scores for all the 8 models. We find the best models as per F-measure, Accuracy and MCC scores. Then we extract explanations using LIME approach and compare the explanations with each of the \textit{Trigger, LoST indicators and Consequences text}. We find that the explanations are more close to the LoST ground truth but still there exist significant similarities with text indicating Triggers and consequences. we need efficient models to shift the attention of NLP models from Trigger and consequences to LoST indicators. }

\begin{table}[!ht]
\centering
% \scriptsize
\small
\caption{Detecting Low Self-Esteem. Comparison of the existing classifiers with Precision (P), Recall (R), F1-score (F1), Accuracy and MCC score, are averaged over 10-fold cross validation. Present: Presence of Low Self-Esteem.}
\label{resultss}
\begin{tabular}{l|p{0.8cm}p{0.8cm}p{0.8cm}|p{0.8cm}|p{0.8cm}}
 \toprule[1.5pt]
\textbf{Model}    & \multicolumn{3}{c}{\textbf{Present}} & \textbf{Acc.} &\textbf{MCC}  \\   & \textbf{P} & \textbf{R} & \textbf{F} & \\ \midrule 
% LSTM &  0.50 & 0.26 & 0.34  & 0.77  & \\
% GRU   & 0.48 & 0.27 & 0.35 & 0.76 \\ 
BERT  &  0.6392 &	0.6889 &	\textbf{0.6631} &	\textbf{0.8552} &	\textbf{0.5717} \\
AlBERT	& \textbf{0.6413} &	0.6556 &	0.6483 &	0.8529 &	0.5554 \\
DistilBERT &	0.5978 &	0.6111 & 0.6044 & 0.8345 &	0.4998 \\
DeBERTa	& 0.6095 &	\textbf{0.7111} & 0.6564 &	0.8459 &	0.5606 \\
% XLNet &	\textbf{0.6500} &	\textbf{0.7222} &	\textbf{0.6842} &\textbf{0.8621} &\textbf{0.5976}\\

% RoBERTa  & 0.51 & 0.67 & 0.58 &  \textbf{0.82}\\
% XLNet & 0.65 & 0.59 & \textbf{0.62} &  0.81  \\
\midrule
% ClinicalBERT & 0.57   &   0.61    &  0.59& 0.83\\
% PsychBERT & 0.65 &     0.52  &    0.58&0.84\\
% MentalBERT   &   0.72    & \textbf{ 0.68 }& \textbf{0.86}\\
MentalBERT	& 0.6500 &	\textbf{0.7222} & \textbf{0.6842} & \textbf{0.8621} & 	\textbf{0.5976} \\
ClinicalBERT&	0.5729 &	0.6111 & 0.5914 &	0.8253 &	0.4808 \\
PsychBERT	& \textbf{0.6528} & 0.5222 & 0.5802 & 	0.8437 &	0.4902 \\

\bottomrule[1.5pt]
\end{tabular}

\end{table}

% \newgeometry{margin=1cm} % modify this if you need even more space
% \begin{landscape}

\begin{table*}
\centering
\small
\caption{Comparison of the explanations obtained by LIME with three different types of textual cues annotated as \texttt{Trigger, LoST indicators, and consequences} using ROUGE scores and BLEU scores. ([Black, blue, red] + bold) represents the highest values of [ATS: All Test Samples, TP: True Positives, TN: True Negatives], respectively. Higher the value of LoST indicators and lower ($\downarrow$) the values with Trigger and consequences, better is the classifier.}
\label{table2}
\begin{tabular}{l|l|p{1cm}|p{1cm}|p{1cm}|p{1cm}|p{1cm}|p{1cm}|p{1cm}|p{1cm}|p{1cm}}
 \toprule[1.5pt]

\textbf{Model}  & Evaluation & \multicolumn{3}{c|}{\textsc{\textbf{Trigger ($\downarrow$)}}}  & \multicolumn{3}{c|}{\textsc{\textbf{LoST Indicators ($\uparrow$)}}}  & \multicolumn{3}{c}{\textsc{\textbf{  Consequences ($\downarrow$)}}} \\
& & \textbf{ATS}  &\textbf{TP}   &\textbf{TN}&\textbf{ATS}  &\textbf{TP}   &\textbf{TN}&\textbf{ATS}  &\textbf{TP}   &\textbf{TN}\\

%&\textbf{ATS (R, B)}  &\textbf{TP (R, B)}   &\textbf{TN (R, B)}&\textbf{ATS (R, B)}  &\textbf{TP (R, B)}   &\textbf{TN (R, B)}\\
\midrule
% &\textbf{(R-ATS, B-ATS)}  &\textbf{(R-TP, B-TP)}   &\textbf{(R-TN, B-TN)}&\textbf{(R-ATS, B-ATS)}  &\textbf{(R-TP, B-TP)}   &\textbf{(R-TN, B-TN)}\\

% &\textbf{R-ATS}  &\textbf{R-TP}   &\textbf{R-TN}&\textbf{R-ATS}  &\textbf{R-TP}   &\textbf{R-TN}\\%  &\textbf{R-ATS}   &\textbf{R-TP} & \textbf{B-TP}  &\textbf{R-TN} & \textbf{B-TN}&\textbf{R-ATS} & \textbf{B-ATS}  &\textbf{R-TP} & \textbf{B-TP}  &\textbf{R-TN} & \textbf{B-TN} \\
% \midrule

% \bottomrule[1.5pt]
% \midrule
% & \textbf{B-ATS} &  \textbf{B-TP} & \textbf{B-TN} & \textbf{B-ATS} &  \textbf{B-TP} & \textbf{B-TN} & \textbf{B-ATS} &  \textbf{B-TP} & \textbf{B-TN} \\
BERT & ROUGE &  0.0756 & 0.0705 &0.0868 & \textbf{0.2773} & 	\textcolor{blue}{\textbf{0.3128}}  &	0.1985	& 0.0421 & 	 0.0442 &0.0375\\
 & BLEU & 0.0519 &	0.0446 &	0.0681&	\textbf{0.1656}& \textcolor{blue}{\textbf{0.1760}}&	0.1426  & 0.0315 &	\textcolor{blue}{\textbf{0.0349}}   &0.0241 \\

% , 0.052) & (	0.044)  & ( 0.068) & (\textbf{0.277}, \textbf{0.166}) & 	(\textcolor{blue}{\textbf{0.313}}, \textcolor{blue}{\textbf{0.176}})  &	(0.198, 0.143) 	& (0.042, 0.031)   & 	 (0.044, \textcolor{blue}{\textbf{0.035}})  & (0.037, 0.024) \\
% & \hline\\
ALBERT & ROUGE & \textbf{0.0987} & \textcolor{blue}{\textbf{0.0973}} & \textcolor{red}{\textbf{0.1014}} & 0.2201 & 0.2424 & 0.1777 & 0.0359 & 0.0427 & 0.0229 \\

& BLEU & \textbf{0.0657} & \textcolor{blue}{\textbf{0.0595}} & \textcolor{red}{\textbf{0.0774}} & 0.1363 & 0.1367 & 0.1356 & 0.0271 & 0.0318 & 0.0179  \\
DistilBERT & ROUGE & 0.0826 & 0.0837 & 0.0809 & 0.2471 & 0.2738 & 0.2051 & 0.0351 & 0.0273 & 0.0474\\
& BLEU & 0.0569 & 0.0564 & 0.0579 & 0.1451 & 0.1412 & 0.1513 & 0.0260 & 0.0219 & 0.0324  \\
DeBERTa & ROUGE & 0.0687 & 0.0645 & 0.0790 & 0.2461 & 0.2570 & \textcolor{red}{\textbf{0.2193} }& \textbf{0.0468} & \textcolor{blue}{\textbf{0.0465}} & \textcolor{red}{\textbf{0.0476}}\\
& BLEU & 0.0459 & 0.0389 & 0.0634 & 0.1586 & 0.1513 & \textcolor{red}{\textbf{0.1724}} & \textbf{0.0346} & 0.0341 & \textcolor{red}{\textbf{0.0359}}\\
\midrule
ClinicalBERT & ROUGE & 0.0728 & 0.0729 & 0.0726 & 0.2045 & 0.2304 & 0.1639 & \textbf{0.0369} & \textcolor{blue}{\textbf{0.0359}} & 0.0387\\
& BLEU & 0.0463 & \textcolor{blue}{\textbf{0.0451}} & 0.0485 & 0.1260 & 0.1272 & 0.1241 & \textbf{0.0286} & \textcolor{blue}{\textbf{0.0288}} & 0.0284 \\
PsychBERT & ROUGE & \textbf{0.0946} & 0.0729 & \textcolor{red}{\textbf{0.1272}} & 0.2310 & 0.2304 & 0.1949 & 0.0307 &  \textcolor{blue}{\textbf{0.0359}} & \textcolor{red}{\textbf{0.0415}}  \\
& BLEU & \textbf{0.0648} & \textcolor{blue}{\textbf{0.0451}} & \textcolor{red}{\textbf{0.0860}} & 0.1336 & 0.1272 & 0.1414 & 0.0226 & \textcolor{blue}{\textbf{0.0288}} & \textcolor{red}{\textbf{0.0325}}\\
MentalBERT & ROUGE & 0.0820 & \textcolor{blue}{\textbf{0.0733}} & 0.1047 & \textbf{0.2614} & \textcolor{blue}{\textbf{0.2821}}  & \textcolor{red}{\textbf{0.2076}} & 0.0160 & 0.0126 & 0.0248 \\
& BLEU & 0.0464 & 0.0348 & 0.0768 & \textbf{0.1544} & \textcolor{blue}{\textbf{0.1521}}  & \textcolor{red}{\textbf{0.1604}}  & 0.0107 & 0.0091 & 0.0147 \\
 \toprule[1.5pt]
\end{tabular}
\end{table*}

\paragraph{Evaluation of Classifiers} Table 2 compares the performance of all the existing classifiers. The domain-specific PLMs, MentalBERT~\cite{ji2022mentalbert} shows the best performance with the highest scores across almost all evaluation metrics (with second best results for Precision after PsychBERT~\cite{vajre2021psychbert}), suggesting more accurate classification. This happens probably due to the domain-specific pre-training of the model on mental health-related posts collected from Reddit. Among the PLMs, BERT demonstrates the highest accuracy (0.8552) followed by AlBERT (0.8529). 
DistilBERT has the lowest scores for all the evaluation metrics, minimally contributing towards classifying texts with low self-esteem. 
%DeBERTa shows a higher recall (0.7111) compared to the most models, which suggests it identifies the majority of low self-esteem instances effectively. However, its precision is relatively lower, indicating a higher rate of false positives. 
ClinicalBERT~\cite{huang2019clinicalbert} has the lowest F1-score (0.5914) and MCC (0.4808) because it is trained on clinical notes, resulting in indifferent nature of the text which is not suitable for our task.
%. Its recall equals AlBERT's, but its precision is the lowest. Similarly, PsychBERT has the highest precision (0.6528), suggesting it can accurately predict a positive case. However, it has the lowest recall score (0.5222), which probably shows that it might miss a portion of actual positive cases. The F1-score, accuracy, and MCC for PsychBERT also fall within the higher performing range among the compared models.  
%ClinicalBERT is 
As such, the MentalBERT outperforms all other models followed by the BERT model, especially in terms of F-score, Accuracy and MCC score, suggesting clear interpretation of the classifier, overall correctness of the classifier, and efficiency of the model in-case of imbalanced dataset, respectively.

\subsubsection{Evaluation of Explanations}
We further evaluate the text-spans focused by attention mechanism to detect low self-esteem. Table 3 offers an in-depth look into the system-level explainability of various NLP models. Through this analysis, we aim to uncover how each classifier deciphers textual cues within a corpus. Using the LIME approach, we extract explanations and gauge their alignment with the three textual cues: \textit{Trigger, LoST indicators}, and \textit{Consequences} by leveraging two similarity metrics: ROUGE and BLEU. A noticeable trend is the superior performance of classifiers that closely align their system-level explanations with the ``LoST indicators.'' This implies a fundamental shift in understanding classifier behavior: models that prioritize recognizing indicators related to LoST generally outdo those emphasizing ``Trigger'' or ``Consequences''. BERT and MentalBERT are exemplary in this regard, leading among the pre-trained Language Models (PLMs) and domain-specific pre-trained language models, respectively. However, models like AlBERT and PsychBERT which exhibited top scores in \textit{Trigger}, and DeBERTa and ClinicalBERT which did well in \textit{Consequences}, underscore a potential inefficiency. Ideally, these models should prioritize \textit{LoST indicators} as key discriminative cues, but instead, they seem to focus on possibly less relevant textual signals. Further granulating our observations, we note that for \textit{LoST indicators}, scores are consistently higher for True Positives (TP) than True Negatives (TN). This indicates a robust attention mechanism, particularly for TP instances, reaffirming that the models discern crucial cues correctly when classifying true instances.

While the classifiers show a promising tilt towards \textit{LoST}, there's still significant alignment with \textit{Triggers} and \textit{Consequences}. This can make explanations unclear, causing confusion and leading to potential misinterpretations. The findings underline a significant challenge: creating models that are better calibrated towards specific indicators, like \textit{LoST}, instead of general cues like \textit{Triggers} or \textit{Consequences}. The latter may be more abundant or explicit in texts but may not always be the most crucial from a psychological perspective.  Future research should explore mechanisms to prioritize \textit{LoST indicators} during model training. This could involve novel attention mechanisms, penalization techniques during training to reduce emphasis on \textit{Triggers} and \textit{Consequences}, or even dataset augmentation to include more explicit \textit{LoST indicators} instances. As NLP models delve deeper into psychological text analysis, their explainability becomes paramount. It's crucial not just to have accurate models but also ones that can clearly and intuitively explain their decision-making rationale, especially in sensitive areas like psychological analysis. Domain-specific PLMs like MentalBERT suggest the utility of fine-tuning on specialized domain-specific corpora. Further explorations could involve curating more granular psychological datasets and refining models on them to better capture nuanced indicators.
%In future, we plan to develop enhancing the attention mechanism for a better classifier by considering such observations. 

%Our experimental results results show that in \textit{Grouping}, MentalBERT ranks true negatives the best, as shown by the highest NDCG score (0.3586). AlBERT performs better overall, leading in Rouge-1 and Bleu-1 scores. In the \textit{Trigger }scenario, we observe that BERT performs best, which shows its performance in handling the n-gram overlap in true positives (Rouge-1 score: 0.3128) and ranking true negatives (NDCG score: 0.3127). We also observe that in the \textit{LoST} \textit{Consequence} scenario, BERT performs better overall  (NDCG scores: 0.2820 and 0.3089, respectively), while MentalBERT performs best in ranking true negatives (NDCG score: 0.4384). Overall, these results show that MentalBERT performs better in ranking true negatives in \textit{Grouping }and \textit{LoST Consequences}, while BERT performs better in \textit{Trigger }and in overall/True Positives ranking in \textit{LoST Consequences}. AlBERT peforms better than others in matching n-grams in \textit{Grouping} as shown in results. To sum up, we find that the explanations are more close to the \textit{LoST} ground truth but still there exist significant similarities with text indicating \textit{Triggers} and \textit{Consequences}. In future, we may need more efficient models to shift the attention of these NLP models from \textit{Trigger} and\textit{ Consequences} to \textit{LoST }indicators

\paragraph{Out of Distribution Analysis}
We test the models on OOD dataset. We carry out the annotation task using the same annotation scheme on the 200 samples of existing dataset and examine classification method in Table~\ref{tab4}. From the general-purpose models, BERT emerged as top performer in terms of F1-score, achieving 0.6667. The balance between precision and recall for BERT demonstrates its capacity to identify low self-esteem indications reliably without raising too many false alarms. DeBERTa's numbers echo this sentiment, with even a slight improvement in accuracy. MentalBERT, as expected from a domain-focused model, exhibited remarkable precision at 0.9231, suggesting that when it flags a text-span as indicative of low self-esteem, it's likely correct. However, its recall sits at 0.4800, hinting that it might miss out on some relevant instances. This reinforces the idea that while domain-specific models might be more accurate in their detections, ensuring reliability in terms of comprehensiveness remains a challenge. Metrics like the MCC play a pivotal role in understanding a model's reliability. For instance, BERT's MCC score of 0.6295 indicates a good quality binary classification, while models with lower MCC scores, like ClinicalBERT at 0.4280, might not be as reliable in distinguishing between the positive and negative classes. The results underscore the fact that while several models can detect low self-esteem indications with decent accuracy, achieving reliable and consistent results is a nuanced challenge. It's not just about flagging potential indicators; it's about ensuring that these detections are genuine, consistent, and that potential indicators aren't overlooked. While BERT present promise, there's room for improvement, especially in the balance between precision and recall. As the stakes involve individuals' psychological well-being, a miss or false alarm isn't trivial. Thus, the quest for a model that can reliably analyze and flag text-spans indicative of low self-esteem on social media platforms, given the ever-evolving nature of online discourse, remains an ongoing challenge.

\begin{table}[!ht]
\centering
\small
\caption{Identification of Low Self-Esteem with \textsc{LoST.v2} dataset. Comparison of the existing classifiers with Precision (P), Recall (R), F1-score (F1), Accuracy and MCC score, are averaged over 10-fold cross validation. Present: Presence of Low Self-Esteem.}
\label{tab4}
\begin{tabular}{l|ccc|c|c}
 \toprule[1.5pt]
\textbf{Model}    & \multicolumn{3}{c}{\textbf{Present}} & \textbf{Acc.} &\textbf{MCC}  \\   & \textbf{P} & \textbf{R} & \textbf{F} & \\ \midrule 
% LSTM &  0.50 & 0.26 & 0.34  & 0.77  & \\
% GRU   & 0.48 & 0.27 & 0.35 & 0.76 \\ 
BERT  &  0.7500 &	0.6000 &	\textbf{0.6667} &	\textbf{0.9242} &	\textbf{0.6295} \\
AlBERT	& 0.8462 &	0.4400 &	0.5789 &	0.9192 &	0.5745 \\
DistilBERT &	0.7500 &	0.3600 & 0.4865 & 0.9040 &	0.4770 \\
DeBERTa	& 0.8235 &	0.5600 & \textbf{\textit{0.6316}} & \textbf{\textit{0.9192}}&	0.6134 \\

\midrule

MentalBERT	& 0.9231 &	0.4800 & \textbf{0.6316} & \textbf{0.9293} & 	\textbf{0.6360}\\
ClinicalBERT&	0.7778 &	0.2800 & 0.4118 &	0.8990 &	0.4280 \\
PsychBERT	& 0.6842& 0.5200 & 0.5909 & 	0.9091 &	0.5473 \\

\bottomrule[1.5pt]
\end{tabular}

\end{table}

\begin{table}
\small
\caption{Observations with the Out-of-distribution (OOD) dataset. Comparison of the explanations obtained by LIME with three different types of textual cues annotated as \texttt{Trigger, LoST indicators, and consequences} using ROUGE scores and BLEU scores. (blue and red) represents the highest values of LoST in True Positives and comparable values of LoST and Trigger, respectively. }
\label{tab5}
\begin{tabular}{l|l|p{1cm}|p{1cm}|p{1cm}}
 \toprule[1.5pt]

\textbf{Model}  & \textbf{Eval.} & \textsc{\textbf{T}}  & \textsc{\textbf{L}}  & \textsc{\textbf{C}} \\
% & & \textbf{TP}  & \textbf{TP} & \textbf{TP} & \textbf{TP} & \textbf{TP} \\

\midrule

BERT & ROUGE & 0.0449 & \textcolor{blue}{0.1764} & 0.0533 \\
& BLEU & 0.0225 & \textcolor{blue}{0.1328} & 0.0432 \\

ALBERT & ROUGE & \textcolor{red}{0.1525} & \textcolor{red}{0.1857} & 0.0259 \\
& BLEU & \textcolor{red}{0.1142} & \textcolor{red}{0.1468} & 0.0166 \\

DistilBERT & ROUGE & 0.0569 & 0.1291 & 0.0266 \\
& BLEU & 0.0138 & 0.0993 & 0.0200 \\

DeBERTa & ROUGE & \textcolor{red}{0.0979} & \textcolor{red}{0.1094} & 0.0159 \\
& BLEU & \textcolor{red}{0.0638} & \textcolor{red}{0.0739} & 0.0100 \\
\midrule
ClinicalBERT & ROUGE & 0.0607 & 0.0968 & 0.0533 \\
& BLEU & 0.0279 & 0.0536 & 0.0500 \\

PsychBERT & ROUGE & 0.1444 & \textcolor{blue}{0.2310} & 0.0352 \\
& BLEU & 0.1018 & \textcolor{blue}{0.1843} & 0.0260 \\

MentalBERT & ROUGE & 0.1054 & \textcolor{blue}{0.2228} & 0.0214 \\
& BLEU & 0.0633 & \textcolor{blue}{0.1897} & 0.0133 \\

 \toprule[1.5pt]
\end{tabular}
\end{table}

An OOD dataset provides an essential playground to test the robustness of classifiers. It's crucial to see how well models generalize and what kind of explanations they provide when confronted with data that is not part of the original dataset. Table~\ref{tab5} captures how the widely acknowledged explanation framework, LIME, explains the models' decisions using ROUGE and BLEU scores. ALBERT and DistilBERT’s explanations, as indicated by its scores, appear to be somewhat skewed towards \textit{Trigger}. Thus, ALBERT and DistilBERT might sometimes conflate triggers or consequences with actual indicators of low self-esteem. DeBERTa's scores are interesting. Its explanations tend to blur the lines between triggers, and indicators. The difference in scores is not significant, implying that DeBERTa might have difficulty distinctly recognizing each type of textual cue. ClinicalBERT's scores reiterate the challenge of distinctly identifying textual cues. While its performance on recognizing LoST indicators is commendable, it seems to also give substantial weight to consequences, as seen from its relatively high BLEU score. PsychBERT stands out, with its ROUGE score for LoST indicators (0.2310) being notably higher than the scores for triggers and consequences. This suggests that PsychBERT's explanations are well-aligned with recognizing actual indicators of low self-esteem. Lastly, MentalBERT, another domain-specific model, also shows a notable distinction in scores, particularly with a higher emphasis on LoST indicators, which is encouraging. It indicates that MentalBERT, in most cases, recognizes and emphasizes the actual indicators over triggers and consequences. However, the performance of MentalBERT is compromised as compare to BERT due to 2.5 times more focus on \textit{Trigger}.

\subsection{Discussion}
In the realm of NLP, we benchmark traditional classifiers to evaluate the performance. However, to enhance the scope of our research, we focus our attention on two paramount models: \textit{BERT} and \textit{MentalBERT}.

\paragraph{Performance.}
With reference to the \textit{ROUGE} score, a widely accepted standard for evaluating text-based tasks, our results were elucidating. The ratio of \textit{[LoST indicators]} with \textit{[Trigger + consequences]} together yield intriguing results. \textit{BERT} registered a ratio of approximately $2.354$, while \textit{MentalBERT} showcased a superior ratio of approximately $2.667$. This distinction in performance is further mirrored when considering the \textit{BLEU} scores. BERT posted a ratio of roughly $1.987$, while MentalBERT notably exceeded this with a ratio close to $2.704$. Drawing from these numerical evaluations, it is evident that, based on our empirical data and selected metrics, \textit{MentalBERT} emerges as the more robust model in comparison to the conventional \textit{BERT} framework. However, \textit{MentalBERT} compromise the performance on OOD data, suggesting \textit{BERT} as more reliable model than \textit{MentalBERT}. We plan to examine the robustness and trustworthiness of two models in the future work. 

\paragraph{Attention Mechanism Error Analysis.}
Notably, the inherent attention mechanism in these classifiers displayed tendencies to stray from concentrating on the \textit{LoST indicators}. This diversion is significant, constituting between $30\%$ and $50\%$ of the textual cues vital for crafting system-level explanations. To address these observed limitations and to steer the discipline towards a more promising direction, we have curated the our dataset for reliability analysis and OOD testing.

\paragraph{Introducing a new dataset.}
 Our dataset is emblematic of a broader vision: urging the scholarly community to prioritize models accentuating trustworthiness, safety, and, most importantly, reliability. While accuracy indisputably remains a pivotal metric, our research underscores the pressing need to transcend the enticements of mere accuracy offered by opaque "black-box" NLP models and to champion transparency and intelligibility.

% Ensuring the consistency and reliability of annotations within an open-source dataset is crucial for comprehending its nature. The systematic categorization and delineation of three distinct types of textual spans play a significant role in aligning models with the task of identifying low self-esteem. The introduction of our newly constructed dataset, paves the way for evaluating the reliability and coherence of models designed for concept extraction and classification in user-penned texts.

\paragraph{Significance of Annotation Consistency.}

The integrity of any open-source dataset largely hinges on the consistency and reliability of its annotations. When working with datasets, especially in the realm of NLP, ensuring that each data point is annotated with precision and uniformity is paramount. This not only serves to validate the authenticity and reliability of the dataset but also offers researchers and model designers a clear comprehension of the underlying structure and nature of the data. As such, achieving consistency in annotations becomes an indispensable step for avoiding biases, ensuring replicability, and ultimately obtaining robust results across various applications and analyses.

\paragraph{Role of Textual Span Categorization.} The systematic categorization and demarcation of textual spans into three distinct types plays a pivotal role in guiding models towards the intricate task of identifying instances of low self-esteem in texts. By providing a clear framework for these spans, models are endowed with the necessary guidance to discern and recognize the subtle nuances and patterns indicative of low self-esteem. 

% \paragraph{Potential of the New Dataset.} With the unveiling of our novel dataset, we are poised at the brink of a transformative phase in the field. This dataset not only stands as a testament to rigorous data collection and curation processes but also acts as an evaluative benchmark for models specifically tailored for the tasks of concept extraction and classification in user-generated texts. Given the ubiquity and diversity of user-penned texts in the digital age.

\paragraph{Ethics and Broader Impact.}
Our dedication lies in upholding ethical principles to safeguard user privacy and anonymity~\cite{henderson2018ethical}. To prevent any misuse, the examples presented in this paper are modified through obfuscation, and paraphrasing. In order to uphold the ethical principles of privacy, safety, and accountability, we have abstained from disclosing any metadata in the public domain. Due to the subjective nature of our task, there may be some inherent biases in our annotations~\cite{zirikly2022explaining}. As we consider explainability as the decision-making parameter, we encourage the enhancement of classifier's attention mechanism in the near future. We design our dataset to facilitate the automated and reliable annotations in identifying various aspects of mental disturbance within a given text~\cite{meyer2022we,wang2021want}. The practical application of this NLP-centered task is the pre-screening of social media users during in-person session of mental health triaging, clinical diagnostic interviewing and motivational interviewing~\cite{artificialintelligencenewsBabylonHealth,westra2011extending}. Moreover, this task elicits both risk and resilience features when monitoring cognitive decline and severe mental disorders. Another practical considerations is its applicability to problems with work-life balance, abusive relationships and impact of job-layoffs during economic recession~\cite{heron2022female,howard2022work}. We acknowledge the need of its licensed use by clinicians, practitioners and other stakeholders to avoid any potential misuse or societal impact of our work.  %. For instance, the research community witnesses a surge in the recent issue of job layoffs since COVID-19 pandemic. 

\paragraph{Limitations.}
While BERT-based architectures are known for their accuracy, they remain largely "black-box" in nature. Despite using evaluation metrics to understand their performance, the underlying reasons for specific classifications are not always transparent, which can be crucial when dealing with sensitive topics like mental health. The performance of models like BERT and its variants largely depends on the quality and quantity of training data. If the dataset does not comprehensively represent the diversity of expressions of low self-esteem across various demographic and socio-cultural groups, the models' generalization capabilities may be limited. Models such as PsychBERT, MentalBERT, and ClinicalBERT, while fine-tuned for specific psychological task in mental health domain, might still miss subtle textual-cues. 
While it's essential to assess the models' performance in unseen data, such evaluations can sometimes lead to results that don't accurately reflect real-world applicability. Thus, OOD dataset evaluation presents its challenges. 

\section{Conclusion}
As the NLP research community progresses towards developing system-level explainable classifiers, our corpus will play a crucial role in advancing the field of information retrieval in the near future. Our task of constructing an advanced corpus, reveals that the classifiers have shown an increased focus on textual cues that emphasize low self-esteem in Reddit posts and emphasise the pressing need of reliability and robust models for healthcare utilization. The annotated explanations in dataset show a closer alignment with the \texttt{LoST indicators}, although significant similarities still exist with the textual cues indicating triggers and consequences. We establish the BERT model trained over the $T_{LoST}$ text-spans for attention and $T_{lse}$ for classification mechanism, as baseline. We further test its reliability using LIME for extracting explanations or focused text-spans by PLM's and testing over OOD dataset to examine the robustness of the classifiers. In future, it would be interesting to develop efficient models that redirects the attention of NLP models from \textit{triggers} and \textit{consequences} towards \textit{LoST indicators} by infusing external knowledge such as domain-specific knowledge graph and commonsense knowledge. %We plan to enhance this dataset with more samples and additional aspects of mental health in social media posts. 

\section*{Acknowledgement}
This project was supported by [Anonymous].
% We express our gratitude to our experts Veena Krishnan and Ruchi Joshi for their unwavering support throughout the project. This project was partially supported by NIH R01 AG068007.

\bibliography{icwsm}

\end{document}